\documentclass[11pt]{article}
\usepackage[utf8]{inputenc}
\usepackage[margin=1in]{geometry}
\usepackage{graphicx}
\usepackage{amsmath, amssymb}
\usepackage{array}
\usepackage{booktabs}
\usepackage{tabularx}
\usepackage{threeparttable}

\usepackage{xcolor}

\definecolor{colorSeg}{HTML}{2A9D8F}    
\definecolor{colorCls}{HTML}{6C5CE7}    
\definecolor{colorGraph}{HTML}{E76F51}  
\definecolor{bgColor}{HTML}{F8FAFC}     
\definecolor{darkText}{HTML}{1E293B}    

\title{Uncertainty Quantification Is Indispensable for Reliable Connectome-Based Graph Learning: A Narrative Review and Case Study}

\author{
	Mansooreh Pakravan$^{1,*}$ \\[0.5ex]
	\small $^{1}$Department of Electrical and Computer Engineering, Tarbiat Modares University,\\
	\small Al Ahmad Street, Tehran 111-14115, Iran \\[0.5ex]
	\small $^{*}$Corresponding author: Mansooreh Pakravan, \texttt{mpakravan@modares.ac.ir} \\
	\small ORCID: 0000-0003-3148-4071
}

\date{}

\begin{document}
	
	\maketitle

\begin{abstract}
	While graph neural networks (GNNs) have shown substantial promise in connectome-based diagnostic classification, deterministic models inevitably suppress pipeline-induced noise and model ambiguities, yielding overconfident predictions. Although uncertainty quantification (UQ) is widely adopted in voxel-level segmentation, its role in connectomic graph learning remains largely unaddressed. This paper presents a comprehensive narrative review of UQ frameworks tailored to connectome graph learning alongside an empirical case study demonstrating the perils of uncalibrated predictions. We delineate sources of aleatoric and epistemic uncertainty across neuroimaging pipelines and review prominent UQ paradigms, from Bayesian approximations and ensemble methods to evidential learning and conformal prediction. In our case study, a temporal Graph Attention Network (GAT) trained on dynamic functional connectivity (dFC) matrices from the SUDMEX CONN dataset achieves $80.0\%$ diagnostic accuracy (F1 = $0.794$) for Cocaine Use Disorder. However, a post-hoc uncertainty audit via Monte Carlo dropout reveals severe overconfidence ($\text{ECE} = 0.127$), with misclassified subjects assigned prediction confidences up to $95\%$. This empirical divergence between discrimination and calibration underscores the confidence paradox in deep connectomics. Our findings establish that rigorous UQ, calibration, and selective prediction mechanisms are indispensable for deploying trustworthy graph-based biomarkers in clinical neuroscience.
\end{abstract}

\section{Introduction}

The quest to decode the human connectome has fundamentally shifted from traditional pairwise statistical correlations toward a sophisticated paradigm of geometric deep learning. Because macroscale brain organization is inherently governed by complex structural wiring and time-varying functional interactions, Graph Neural Networks (GNNs) have emerged as the standard mathematical language for computational neuroimaging~\cite{ref_alavi_bibliometric, ref_bessadok_tpami}. By representing discrete anatomical parcels or regions of interest (ROIs) as nodes and their synchronized blood-oxygen-level-dependent (BOLD) dynamics or tractography-derived streamlines as edges, graph-based architectures have demonstrated outstanding discriminatory power. These models have achieved remarkable success across diverse translational objectives, including the discovery of network biomarkers for psychiatric and neurodevelopmental disorders~\cite{ref_alavi_bibliometric}, mapping fine-grained sex-specific dimorphisms across distinct parcellation schemes~\cite{ref_fxpnet, ref_pakravan_sex}, and elucidating the topological reorganization and functional hub degradation associated with chronic pain syndromes~\cite{ref_pakravan_osteo}. Recent methodological extensions into explainable graph architectures and causal attention mechanisms have further advanced our mechanistic understanding, enabling researchers to isolate aberrant white-matter pathways in developmental prosopagnosia~\cite{ref_ergan} and track the dynamic trajectory of narrative comprehension hubs during naturalistic continuous listening~\cite{ref_saket_neuroinfo, ref_node_embedding}.

However, as geometric deep learning transitions from exploratory proof-of-concept benchmarks toward safety-critical clinical deployment, a silent crisis of diagnostic trustworthiness has surfaced. The vast majority of neuroimaging machine learning studies remain exclusively evaluated on aggregate performance metrics, such as classification accuracy, sensitivity, specificity, and the area under the receiver operating characteristic curve (AUC)~\cite{ref_kumar_annals}. Although these population-averaged metrics confirm broad statistical capability, they provide no mathematical guarantee regarding the reliability or confidence of individual, subject-level predictions~\cite{ref_guo_calibration}. In high-stakes neurological and psychiatric assessments, a diagnostic framework reporting an impressive 80\% overall accuracy remains clinically hazardous if its remaining 20\% error margin is accompanied by unyielding, unwarranted statistical certainty~\cite{ref_faghani_radiology}. Modern deep architectures are notorious for producing uncalibrated, overconfident softmax probabilities that diverge dramatically from empirical likelihoods of diagnostic correctness~\cite{ref_guo_calibration, ref_lambert_ai_med}.

This vulnerability is exceptionally pronounced in graph-structured neuroimaging pipelines. Recent theoretical and empirical analyses reveal that standard post-hoc calibration strategies, such as temperature scaling and vector scaling, frequently break down when applied to graph architectures~\cite{hsu2024what}. The iterative message-passing schemes fundamental to GNNs propagate and aggregate localized node representations across topological neighborhoods, which inadvertently homogenizes feature distributions, dampens predictive entropy, and masks localized ambiguities~\cite{hsu2024what}. In the context of whole-brain clinical network classification, such structural miscalibration exacerbates false-positive diagnostics and blinds the model to genuine neuroanatomical atypicalities, posing a direct threat to safe patient stratification~\cite{ren2024calibration}.

Predictive uncertainty in neuroimaging is not a monolithic artifact but a multi-faceted phenomenon arising from distinct mathematical and physical origins. It is broadly partitioned into \textit{aleatoric} uncertainty, reflecting irreducible stochasticity such as physiological noise, head motion artifacts, slice-timing discrepancies, and the profound biological heterogeneity underlying complex brain disorders~\cite{ref_hullermeier_uncertainty, ref_kendall_gal}, and \textit{epistemic} uncertainty, which stems from deficiencies in model knowledge, severe sample-to-feature ratio imbalances ($N \ll P$), and class imbalance~\cite{ref_huang_media, ref_mohammadi_brainsci}. Crucially, epistemic uncertainty becomes dominant under inter-site domain shifts caused by scanner hardware differences, field strengths, gradient non-linearities, and disparate acquisition protocols~\cite{he2023epistemic}. In multi-center functional connectomics, when a predictive model encounters out-of-distribution (OOD) geometries induced by scanner variability, unquantified epistemic uncertainty manifests as silent, highly confident diagnostic failures~\cite{he2023epistemic, ref_vega_lara_2026}.

To counteract these vulnerabilities, specialized uncertainty-aware frameworks have gradually been explored. Initial attempts incorporated fuzzy set theory into graph modeling to capture the intrinsic vagueness and boundary ambiguity of resting-state functional connectivity~\cite{ref_pakravan_fuzzycal}. In particular, Type-II fuzzy GNNs demonstrated that modeling edge-weight imprecision enables the extraction of robust biological signatures that remain obscured within deterministic counterparts~\cite{ref_pakravan_sex}. Similarly, fuzzy causal attention mechanisms have bolstered the stability of psychiatric classifications against noise-induced graph perturbations~\cite{ref_pakravan_fuzzycal}. Parallel efforts within the field of probabilistic machine learning have leveraged Bayesian Neural Networks (BNNs), Monte Carlo (MC) Dropout, Deep Ensembles, Laplace approximations, and Evidential Deep Learning to estimate predictive posterior distributions and capture model hesitation~\cite{ref_kendall_gal, muller2021uncertainty}.

Despite these advances, a striking methodological disparity persists across the neuroimaging domain. While uncertainty quantification (UQ) and spatial calibration protocols have achieved widespread validation in localized voxel-level tasks, such as lesion and brain tumor segmentation~\cite{duran2022calibration, ref_fuchs_pmlr, ref_molchanova_ms}, systematic UQ remains critically underrepresented in connectomics and whole-graph classification~\cite{muller2021uncertainty, ref_mohammadi_brainsci}. Functional connectivity matrices are notoriously susceptible to low signal-to-noise ratios, arbitrary thresholding criteria, and atlas-induced structural biases~\cite{ref_pakravan_osteo}. When connectomic graphs are processed deterministically, upstream measurement errors propagate through iterative graph convolutional layers unchecked, compromising the integrity of clinical decision boundaries~\cite{muller2021uncertainty, ren2024calibration}.

To address this methodological divide and provide rigorous finite-sample error control, the machine learning community has increasingly turned toward Conformal Prediction~\cite{ref_angelopoulos_gentle, ref_neurips_conformal}. Conformal prediction offers a distribution-free, model-agnostic framework that transforms brittle point predictions into statistically valid prediction sets guaranteed to cover the ground truth at a user-specified confidence level ($1-\alpha$)~\cite{gordon2023conformalized}. While the classical exchangeability assumption of conformal inference is routinely violated by the spatial and topological dependencies of brain networks, recent theoretical breakthroughs have established valid node-level and whole-graph-level conformal frameworks capable of handling non-exchangeable topological data~\cite{ref_neurips_conformal, lu2023conformal}. By shifting the diagnostic question from ``which single class is most likely?'' to ``which set of candidate diagnoses contains the true clinical state with provable coverage guarantees?'', conformalized medical imaging establishes a principled safeguard for human-in-the-loop decision-making~\cite{gordon2023conformalized, lu2023conformal}.

This narrative review synthesizes the current state of uncertainty-aware machine learning in neuroscience and neuroimaging. We first introduce a problem-oriented taxonomy of uncertainty, distinguishing aleatoric and epistemic sources across neuroimaging pipelines. We then review the principal estimation paradigms used in the literature, including Bayesian approximations, Monte Carlo dropout, deep ensembles, evidential learning, Laplace-based inference, and conformal prediction. Next, we examine how these ideas have been reflected in current neuroscience applications, with particular attention to segmentation, diagnostic classification, and connectomic analysis. To ground the discussion, the review also presents an empirical case study in dynamic functional connectomics, illustrating how discrimination, calibration, and confidence may diverge in graph-based prediction. Together, these sections highlight the practical need for uncertainty-aware and better-calibrated models in neuroimaging, and they motivate future work on trustworthy inference in clinical neuroscience.

\section{Related Works: Uncertainty in Brain Modelling}

Uncertainty is central to neuroimaging machine learning, capturing both intrinsic data noise and the limitations of model knowledge. Conventional deep models output point estimates (e.g., softmax vectors), which fail to distinguish between typical cases and those that are ambiguous or out-of-distribution (OOD)~\cite{ref_lambert_ai_med, ref_kurz_medical_uq}. A more robust approach shifts from point estimates to predictive distributions $p(y^* \mid x^*, \mathcal{D})$, which quantify both belief and confidence~\cite{ref_blundell_bbb, ref_gal_dropout}.

\subsection{Sources of Uncertainty: Aleatoric and Epistemic}

Uncertainty is commonly decomposed into aleatoric and epistemic components~\cite{ref_kendall_gal}. \textbf{Aleatoric uncertainty} (data-intrinsic) arises from head motion, noise, or biological ambiguity (e.g., overlap between ADHD and ASD). It is irreducible and answers: \textit{``Is this observation inherently noisy?''} \textbf{Epistemic uncertainty} (model-related) reflects limited knowledge, high when data deviates from training patterns (e.g., scanner shifts, rare subtypes). It answers: \textit{``Does the model have sufficient experience with this case?''}

\begin{figure*}[t]
	\centering
	\includegraphics[width=\textwidth]{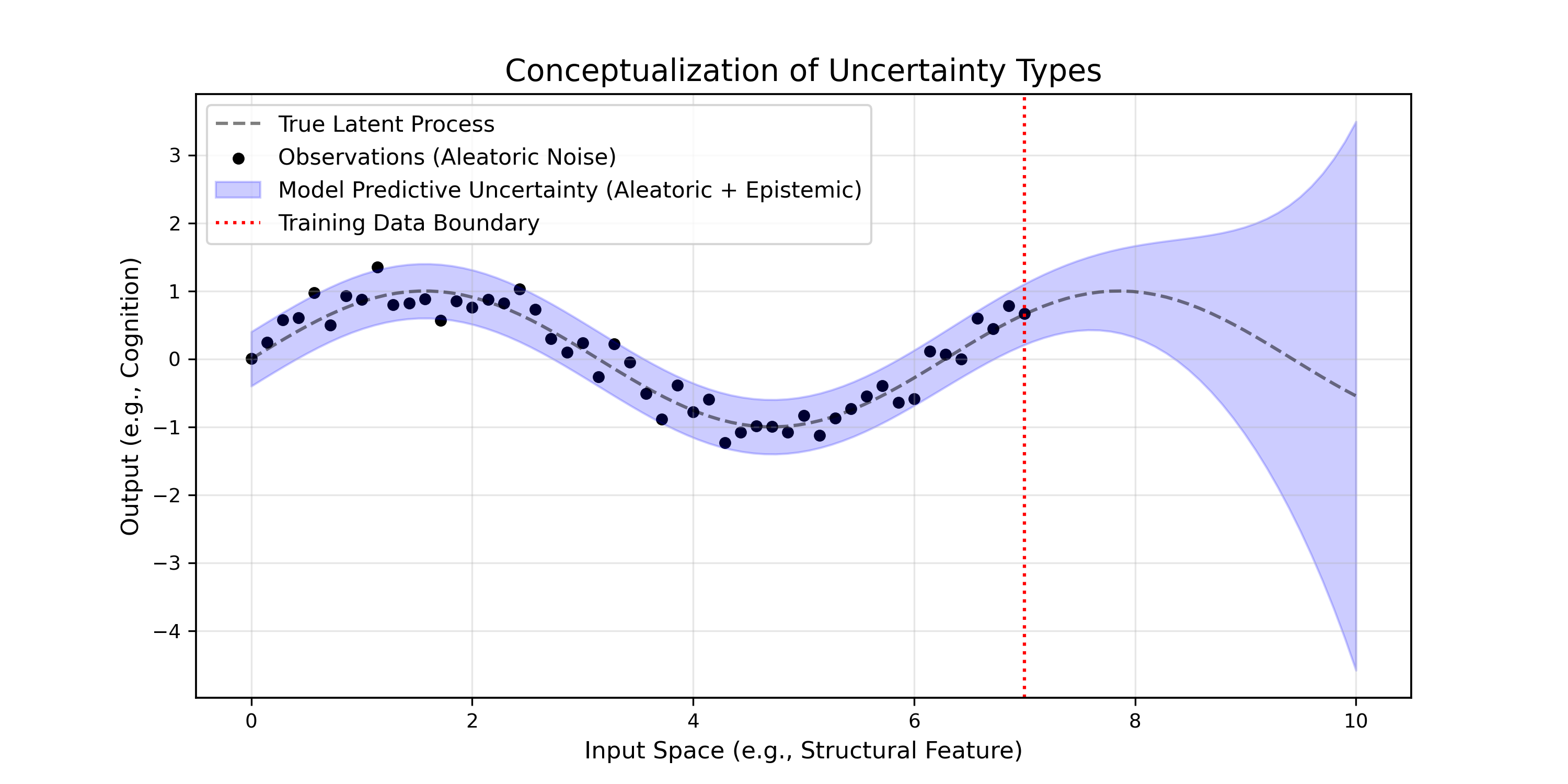}
	\caption{Conceptual simulation of aleatoric and epistemic uncertainty. Persistent noise around the latent process represents aleatoric uncertainty. The widening predictive interval beyond the training boundary ($x=7$) illustrates epistemic uncertainty due to extrapolation.}
	\label{fig:aleatoric_epistemic}
\end{figure*}

These concepts are illustrated in Fig.~\ref{fig:aleatoric_epistemic}. Aleatoric uncertainty persists as scatter around the latent process, while epistemic uncertainty widens predictive intervals beyond the training boundary ($x=7$).

\begin{figure*}[t]
	\centering
	\includegraphics[width=\textwidth]{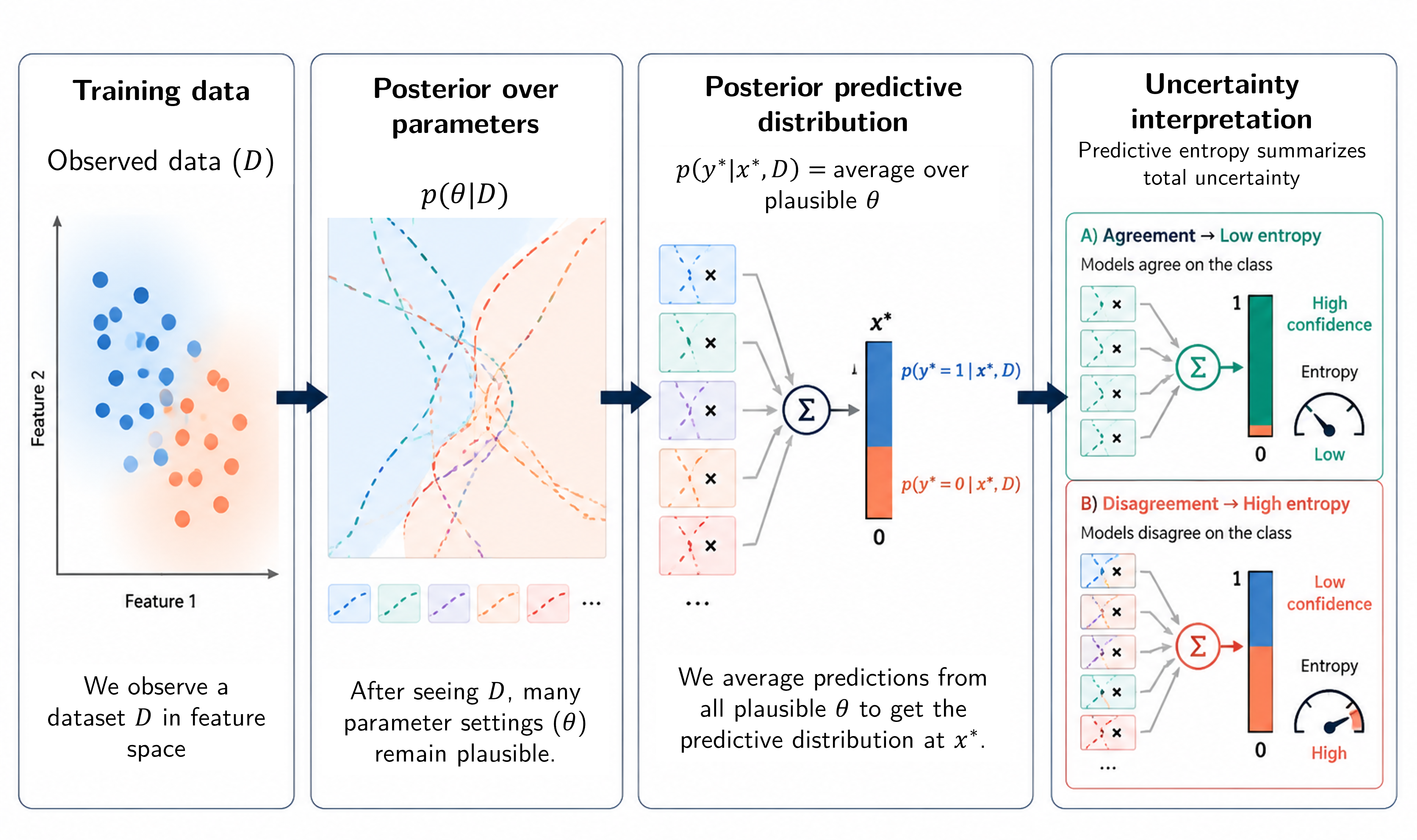}
	\caption{Bayesian view of predictive uncertainty: predictions are averaged over a posterior of plausible parameter configurations $p(\theta|\mathcal{D})$. Model disagreement signifies epistemic uncertainty.}
	\label{fig:bayesian_view}
\end{figure*}

A Bayesian formulation provides a formal framework for this decomposition by averaging predictions over plausible parameters $\theta$:
\begin{equation}
	p(y^* \mid x^*, \mathcal{D}) = \int p(y^* \mid x^*, \theta)\, p(\theta \mid \mathcal{D})\, d\theta.
\end{equation}
Fig.~\ref{fig:bayesian_view} visualizes how agreement among plausible models minimizes epistemic uncertainty, whereas disagreement increases it, serving as a proxy for OOD detection.

\subsection{Methodological Landscape for UQ}

\begin{figure}[t]
	\centering
	\includegraphics[width=\linewidth]{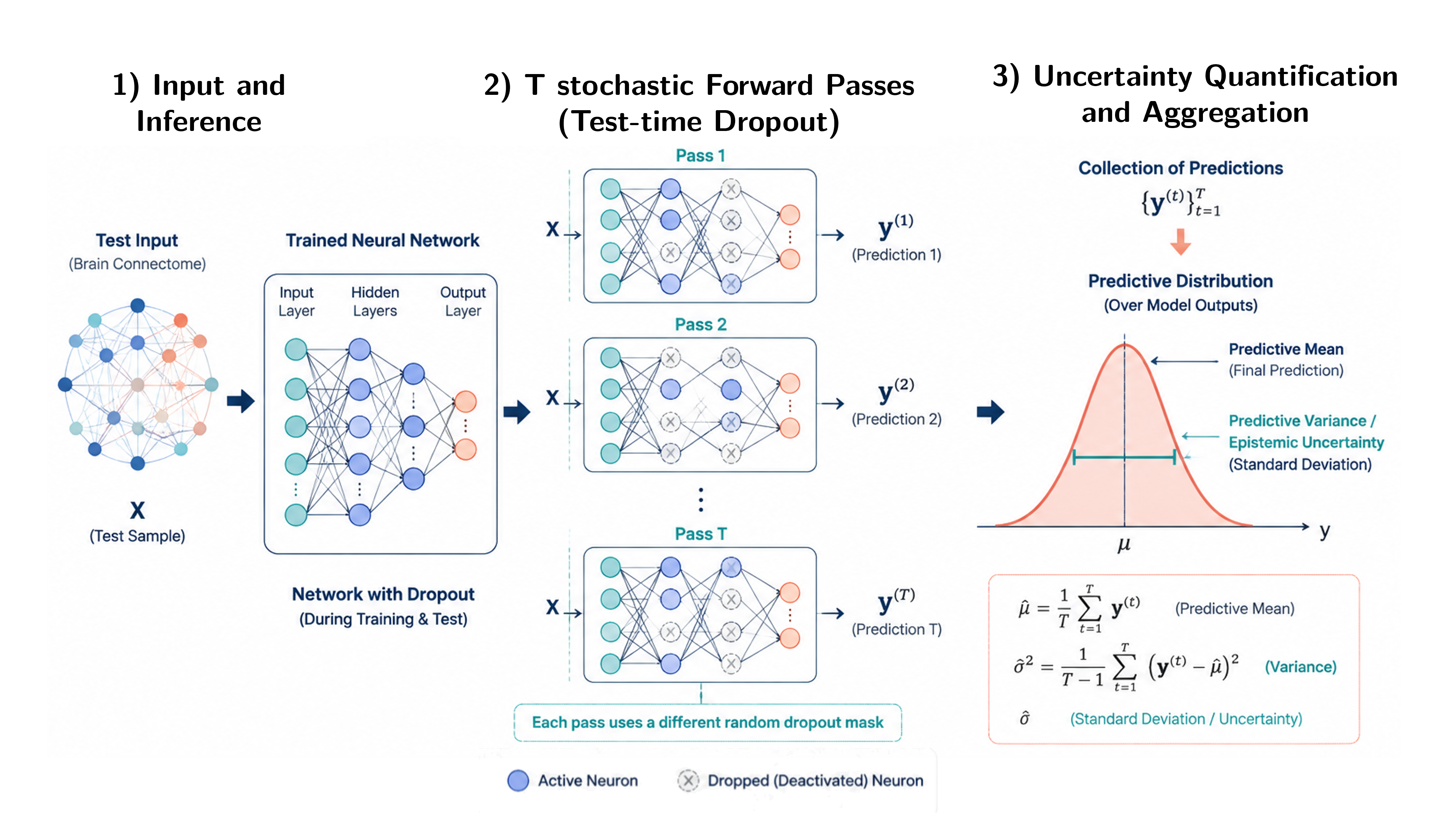}
	\caption{MC dropout estimation: stochastic masks generate an ensemble of sub-networks, where variance quantifies model uncertainty.}
	\label{fig:mc_dropout}
\end{figure}

Various methods address these uncertainties, ranging from parameter-focused approaches to distributional modeling:

\textbf{Epistemic and Ensemble Methods:} Bayesian Neural Networks (BNNs)~\cite{ref_blundell_bbb} and Laplace approximations~\cite{ref_daxberger_laplace} represent parameter uncertainty through distributions. Monte Carlo (MC) dropout~\cite{ref_gal_dropout} offers a computationally efficient proxy by performing stochastic forward passes to estimate model disagreement (Fig.~\ref{fig:mc_dropout}). Deep Ensembles~\cite{ref_lakshminarayanan_ensembles} aggregate independently trained models, providing strong empirical calibration, while SWAG~\cite{maddox_swag} approximates the posterior via gradient trajectory.

\textbf{Evidential and Aleatoric Models:} Evidential Deep Learning (EDL)~\cite{ref_sensoy_edl} models the evidence supporting each class (often via a Dirichlet distribution), allowing for the distinction between low support and conflicting evidence. Conversely, heteroscedastic models explicitly predict input-dependent variance, directly capturing aleatoric uncertainty.

\textbf{Set-based and Vague Representations:} Conformal Prediction (CP)~\cite{ref_angelopoulos_gentle} provides distribution-free prediction sets with rigorous coverage guarantees, moving beyond explicit probability modeling. Similarly, fuzzy systems~\cite{ref_mendel_type2} represent graded membership in overlapping categories, suitable when class boundaries are linguistically vague rather than probabilistically ambiguous.

\subsection{Calibration and Reliability Assessment}
\label{subsec:calibration_reliability}

In high-stakes clinical neuroimaging applications, raw predictive accuracy alone is an insufficient criterion for deployment; model calibration is critical to ensure that predicted confidence bounds faithfully reflect empirical likelihood~\cite{ref_guo_calibration}. Formally, a classifier is perfectly calibrated when the predicted confidence matches its true posterior accuracy:
\begin{equation}
	\mathbb{P}\left(\hat{Y} = Y \;\middle|\; \hat{P} = p\right) = p, \quad \forall p \in [0, 1].
\end{equation}

\begin{figure*}[t]
	\centering
	\includegraphics[width=\textwidth]{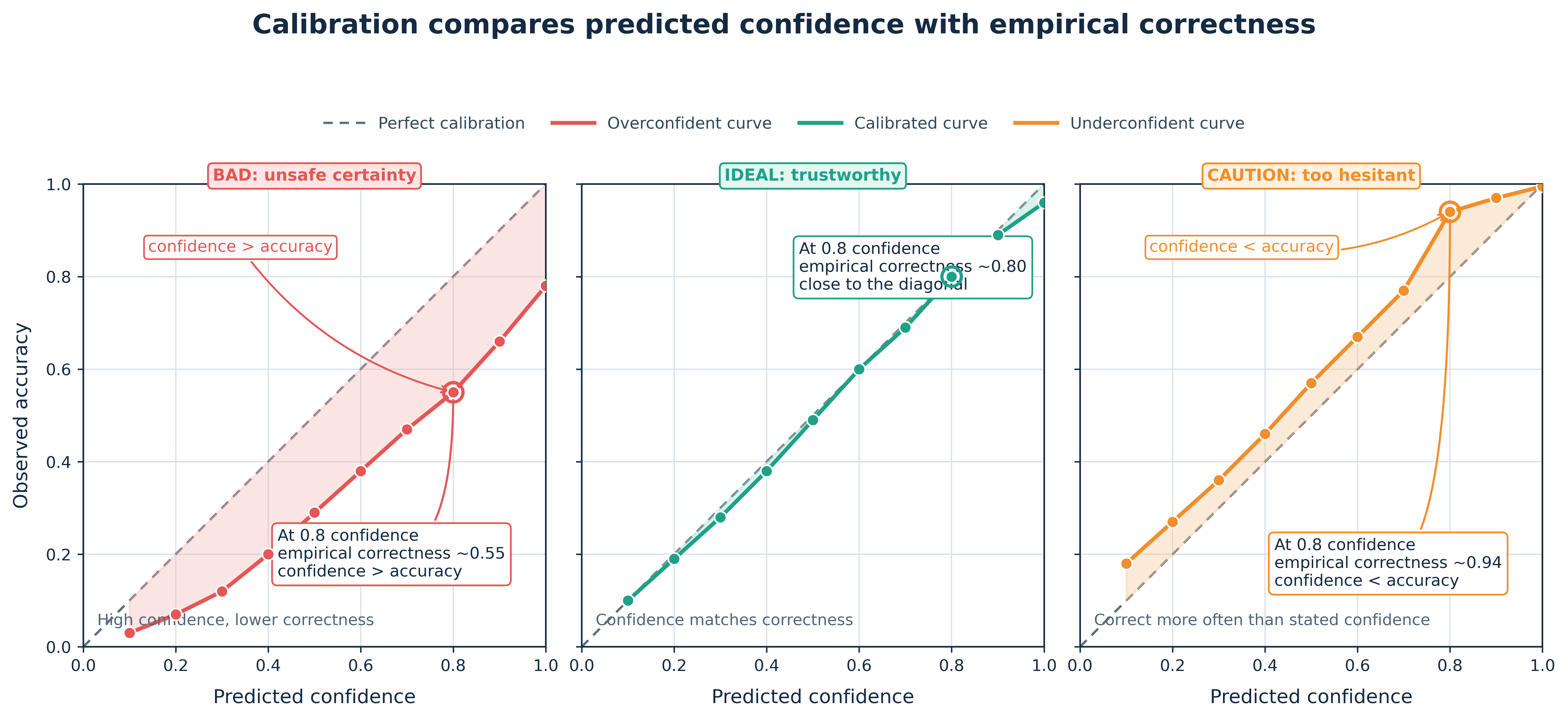}
	\caption{Calibration regimes in predictive modeling: (a) overconfidence, where assigned probabilities outpace true accuracy; (b) ideal calibration, adhering to the diagonal identity; and (c) underconfidence, where the network is more accurate than its output probabilities indicate.}
	\label{fig:calibration_three_panels}
\end{figure*}

Empirical deviations from this ideal alignment fall into distinct calibration regimes, as illustrated in Fig.~\ref{fig:calibration_three_panels}:
\begin{itemize}
	\item \textbf{Overconfidence (Fig.~\ref{fig:calibration_three_panels}a):} The model assigns high subjective confidence that drastically exceeds empirical accuracy ($\hat{P} > \text{acc}$), a pathology pervasive in deep graph architectures.
	\item \textbf{Ideal Calibration (Fig.~\ref{fig:calibration_three_panels}b):} The empirical accuracy curve tracks the diagonal identity line perfectly across all confidence bins.
	\item \textbf{Underconfidence (Fig.~\ref{fig:calibration_three_panels}c):} The model underestimates its predictive reliability ($\hat{P} < \text{acc}$), leading to overly cautious clinical deferrals.
\end{itemize}

\begin{figure}[htp]
	\centering
	\includegraphics[width=\linewidth]{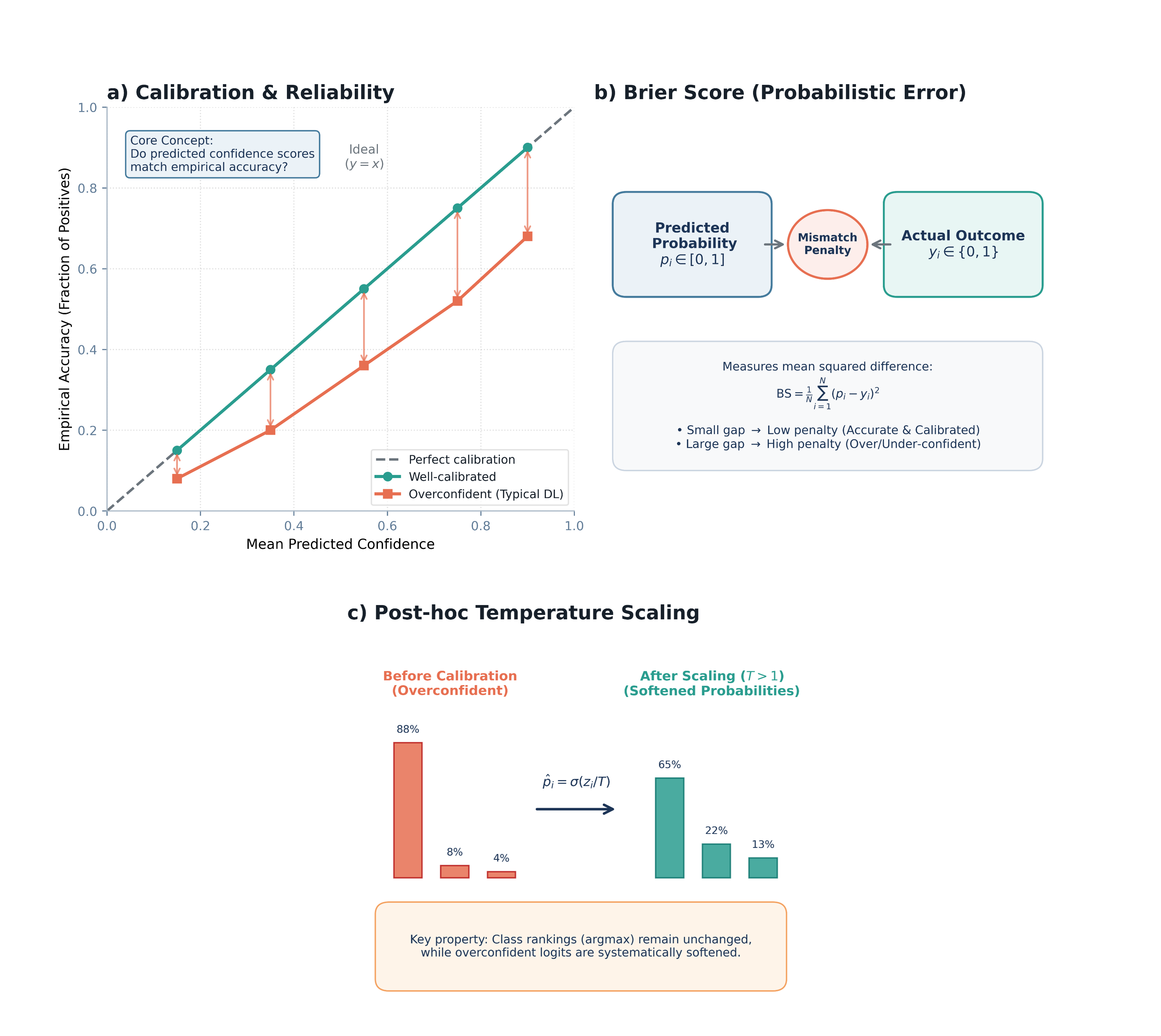}
	\caption{Comprehensive calibration assessment and mitigation framework: (a) Reliability diagram depicting the gap between binned confidence and empirical accuracy (ECE); (b) Brier score formulation for penalty-based probabilistic alignment; and (c) Temperature scaling mechanism for post-hoc logit entropy adjustment.}
	\label{fig:calibration_concept}
\end{figure}

To systematically evaluate and rectify these calibration discrepancies, standard diagnostic and post-hoc mitigation pipelines are employed (Fig.~\ref{fig:calibration_concept}):
\begin{enumerate}
	\item \textbf{Reliability Diagrams and ECE (Fig.~\ref{fig:calibration_concept}a):} Grouping predictions into $M$ equally spaced confidence intervals $B_m$ allows computing the Expected Calibration Error (ECE):
	\begin{equation}
		\text{ECE} = \sum_{m=1}^M \frac{|B_m|}{N} \left| \text{acc}(B_m) - \text{conf}(B_m) \right|,
	\end{equation}
	which quantifies the weighted absolute deviation from the ideal diagonal.
	\item \textbf{Strict Proper Scoring Rules (Fig.~\ref{fig:calibration_concept}b):} The Brier score assesses both sharpness and probabilistic correctness by measuring the mean squared error between the predictive probability vector $\hat{\mathbf{p}}_i$ and the one-hot ground-truth label $\mathbf{y}_i$:
	\begin{equation}
		\text{BS} = \frac{1}{N} \sum_{i=1}^N \sum_{k=1}^K \left( \hat{p}_{i,k} - y_{i,k} \right)^2.
	\end{equation}
	\item \textbf{Post-hoc Logit Recalibration (Fig.~\ref{fig:calibration_concept}c):} When a network exhibits structural overconfidence, Temperature Scaling~\cite{ref_guo_calibration} softens the unnormalized output logits $\mathbf{z}_i$ via a single learned parameter $T > 0$:
	\begin{equation}
		\hat{p}_{i,k} = \frac{\exp(z_{i,k} / T)}{\sum_{j=1}^K \exp(z_{i,j} / T)}.
	\end{equation}
	By setting $T > 1$, the entropy of the predictive distribution is elevated without altering the argmax classification boundary or requiring costly architectural retraining.
\end{enumerate}

\section{Empirical Case Study: A Cautionary Tale}

\subsection{Study Objective and Analytical Rationale}

To illustrate why predictive performance alone is insufficient for trustworthy
neuroimaging machine learning, we conducted an empirical case study using
subject-specific dynamic functional brain graphs. The central purpose of this
analysis was not to claim that a high-performing classifier is necessarily
unreliable, but to examine whether its confidence estimates were consistent
with its actual predictive correctness. In particular, we investigated the
following apparent paradox: a graph neural network (GNN) may achieve an
accuracy close to 80\% while assigning high predictive uncertainty to a
substantial proportion of individual subjects.

This situation is plausible in neuroimaging because the brain is not organized
according to sharply separated diagnostic categories. Subjects within the same
clinical group may exhibit different patterns of functional dysconnectivity,
whereas subjects from different groups may share partially overlapping network
profiles. Consequently, a classifier may learn discriminative population-level
patterns while remaining uncertain about individual cases. This distinction is
especially important in substance-use and psychiatric neuroimaging, where
biological heterogeneity, measurement noise, motion-related artifacts, and
site-specific effects may coexist~\cite{ref_hullermeier_uncertainty,
	ref_lambert_ai_med}.

\subsection{Dataset and Participant Characteristics}
\label{subsec:dataset_participants}

We used the publicly available SUDMEX CONN dataset, a multimodal
neuroimaging resource developed to investigate substance-use disorders
\cite{ref_sudmex_conn}. The cohort included 138 participants, comprising
74 individuals diagnosed with cocaine use disorder (CUD) and 64
age- and sex-matched healthy controls (HC). Although the database also
contains structural MRI and diffusion-weighted imaging, the present analysis
used only resting-state functional magnetic resonance imaging (rs-fMRI) data
for dynamic functional-connectivity analysis and graph-based classification.

The demographic and clinical characteristics of the analytical cohort are
summarized in Table~\ref{tab:sudmex_characteristics}. The CUD and HC groups
were comparable in age and sex distribution. The mean age was
\(30.60 \pm 8.26\) years in the CUD group and
\(30.99 \pm 7.25\) years in the HC group (\(p=0.42\)). The CUD group included
65 males and 9 females, whereas the HC group included 51 males and 11
females. As expected, the CUD group had a significantly higher mean AMAI
score than the HC group
(\(133.45 \pm 50.63\) versus \(107.96 \pm 50.64\), \(p<0.001\)).
The dataset also included information on substance-use history, psychiatric
status, and psychosocial characteristics, including the Addiction Severity
Index (ASI). These variables were used for cohort characterization and were
not included as graph input features unless explicitly stated otherwise.

\begin{table}[htpb]
	\centering
	\caption{Demographic and clinical characteristics of the SUDMEX CONN cohort.}
	\label{tab:sudmex_characteristics}
	\renewcommand{\arraystretch}{1.25}
	\begin{tabular}{lcc}
		\hline
		\textbf{Characteristic} &
		\textbf{CUD} (\(n=74\)) &
		\textbf{HC} (\(n=64\)) \\
		\hline
		Age, years; mean $\pm$ SD &
		\(30.60 \pm 8.26\) &
		\(30.99 \pm 7.25\) \\
		
		Sex, male/female &
		\(65/9\) &
		\(51/11\) \\
		
		AMAI score; mean $\pm$ SD &
		\(133.45 \pm 50.63\) &
		\(107.96 \pm 50.64\) \\
		\hline
	\end{tabular}
\end{table}

\subsection{rs-fMRI Preprocessing and Anatomical Parcellation}
\label{subsec:fmri_preprocessing}

The rs-fMRI data were preprocessed using the Analysis of Functional
NeuroImages (AFNI) software suite~\cite{ref_afni}. The preprocessing pipeline
included slice-timing correction, rigid-body motion correction, coregistration
of functional images to each participant's anatomical T1-weighted image,
spatial normalization to a common MNI template, and spatial smoothing using a
Gaussian kernel with a full width at half maximum (FWHM) of \(5~\mathrm{mm}\).

The data were then temporally band-pass filtered between
\(0.01\) and \(0.15~\mathrm{Hz}\). Nuisance regression was performed to reduce
the influence of non-neuronal signals using white-matter, cerebrospinal-fluid,
and motion-related regressors, together with their temporal derivatives.
Finally, a brain mask was applied to exclude non-brain voxels.

To reduce the dimensionality of the voxel-wise data, we used the
Harvard--Oxford cortical structural atlas, specifically the
\texttt{cort-maxprob-thr25-2mm} version
\cite{ref_harvard_oxford,ref_afni}. This parcellation contains 48 cortical
regions of interest (ROIs). For each participant, the mean BOLD signal was
extracted from every ROI, resulting in the regional time-series matrix
\begin{equation}
	\mathbf{X}^{(s)} \in \mathbb{R}^{T \times N},
\end{equation}
where \(s\) indexes the participant, \(T=300\) denotes the number of temporal
samples, and \(N=48\) denotes the number of cortical nodes. Each column of
\(\mathbf{X}^{(s)}\) represents time series of one cortical
region.

\subsection{Construction of Dynamic Functional Brain Graphs}
\label{subsec:dynamic_graphs}

To represent temporal changes in functional connectivity, the regional time
series of each participant were divided into \(K\) non-overlapping temporal
windows of length \(L=100\) time points. Given \(T=300\) time points, this
procedure yielded $	K = \left\lfloor \frac{T}{L} \right\rfloor = 3 $
graph snapshots per participant.

For participant \(s\) and temporal window \(k\), we calculated the partial
correlation matrix $\mathbf{P}^{(s,k)}=	\left[p_{ij}^{(s,k)}\right]_{i,j=1}^{N}$,

where \(p_{ij}^{(s,k)}\) estimates the direct linear association between
regions \(i\) and \(j\) after statistically controlling for the remaining
regions. To obtain non-negative edge weights, we used the magnitude of the
partial correlations:
$ a_{ij}^{(s,k)} =	\left|p_{ij}^{(s,k)}\right|, \qquad i \neq j$,

and set $a_{ii}^{(s,k)}=0$.

This transformation preserves the strength of the association but discards
its sign; consequently, positive and negative functional relationships are
treated as equally strong connections, which is a limitation of the adopted
representation.

Because partial-correlation matrices are typically dense, we applied
subject- and window-specific percentile-based sparsification. Let
\(\tau^{(s,k)}\) denote the 75th percentile of the non-zero off-diagonal
connection weights in the corresponding snapshot. The resulting weighted
sparse graph was defined as $G^{(s,k)}=(V,E^{(s,k)})$,
where $(i,j)\in E^{(s,k)} \quad \Longleftrightarrow \quad	a_{ij}^{(s,k)} \geq \tau^{(s,k)}$.

The edge weight \(a_{ij}^{(s,k)}\) was retained for all surviving edges.
Thus, each participant was represented by a sequence of three dynamic graph
snapshots:
$\mathcal{G}^{(s)}=\left\{G^{(s,1)}, G^{(s,2)}, G^{(s,3)}\right\}$.

\subsection{Node-Level Graph Features}
\label{subsec:node_features}

In addition to the connectivity matrices, we extracted two node-level
graph-theoretic features from every snapshot: unweighted node degree and mean
neighbor degree. For an undirected graph \(G=(V,E)\), the degree of node \(v\)
is defined as $	d(v)=\left|	\left\{	u:(u,v)\in E\right\}\right|$.

For the weighted graphs, the corresponding weighted degree, or node strength,
was defined as $s(v) =	\sum_{u\in\mathcal{N}(v)} a_{uv}$,

where \(\mathcal{N}(v)\) denotes the set of neighbors of node \(v\).
Although node strength was available from the weighted adjacency matrix, the
degree-based representation was used as the primary node feature. The mean
neighbor degree was calculated as
\begin{equation}
	d_{\mathrm{nn}}(v)
	=
	\frac{1}{|\mathcal{N}(v)|}
	\sum_{u\in\mathcal{N}(v)}d(u).
\end{equation}

Accordingly, the node-feature matrix for each graph snapshot was represented
as $\mathbf{H}^{(s,k)}	\in	\mathbb{R}^{N\times F},	\qquad F=2$,

where the two columns correspond to \(d(v)\) and
\(d_{\mathrm{nn}}(v)\), respectively. These features provide a compact
description of local connectivity and nodal influence while maintaining a
manageable input dimensionality.

\subsection{Graph Neural Network Classifier}
\label{subsec:gnn_classifier}

The dynamic graph sequence was analyzed using a graph neural network (GNN)
classifier. The first graph-convolutional operation was implemented using a
graph-attention layer. For node \(i\), the updated representation was computed
as
\begin{equation}
	\mathbf{h}'_{i}
	=
	\sigma\left(
	\sum_{j\in\mathcal{N}(i)}
	\alpha_{ij}\mathbf{W}\mathbf{h}_{j}
	\right),
\end{equation}
where \(\mathbf{h}_{i}\in\mathbb{R}^{d}\) is the input feature vector,
\(\mathbf{h}'_{i}\in\mathbb{R}^{d'}\) is the updated feature vector,
\(\mathbf{W}\in\mathbb{R}^{d'\times d}\) is a trainable linear transformation,
and \(\sigma(\cdot)\) is a nonlinear activation function.

The attention coefficient between nodes \(i\) and \(j\) was defined as
\begin{equation}
	\alpha_{ij}
	=
	\frac{
		\exp\left(
		\operatorname{LeakyReLU}
		\left(
		\mathbf{a}^{\mathsf{T}}
		\left[\mathbf{W}\mathbf{h}_{i}\,\Vert\,\mathbf{W}\mathbf{h}_{j}\right]
		\right)
		\right)
	}{
		\displaystyle\sum_{\ell \in \mathcal{N}(i)}
		\exp\left(
		\operatorname{LeakyReLU}
		\left(
		\mathbf{a}^{\mathsf{T}}
		\left[\mathbf{W}\mathbf{h}_{i}\,\Vert\,\mathbf{W}\mathbf{h}_{\ell}\right]
		\right)
		\right)
	}
\end{equation}

where \(\Vert\) denotes vector concatenation and
\(\mathbf{a}\in\mathbb{R}^{2d'}\) is a learnable attention vector. This
mechanism enables the model to assign different importance to neighboring
brain regions rather than treating all functional connections identically.

The node-level representations were subsequently aggregated into a
graph-level representation for each temporal snapshot. Let
\(\operatorname{READOUT}(\cdot)\) denote the graph-pooling operation. The
snapshot-level embedding was defined as
$
\mathbf{z}^{(s,k)}
=
\operatorname{READOUT}
\left(
\left\{
\mathbf{h}^{(s,k)}_{i}
\right\}_{i=1}^{N}
\right).
$

The embeddings from the three temporal windows were then combined to form a
participant-level representation:
$
\mathbf{z}^{(s)}
=
\operatorname{AGGREGATE}
\left(
\mathbf{z}^{(s,1)},
\mathbf{z}^{(s,2)},
\mathbf{z}^{(s,3)}
\right).
$

Finally, a fully connected classification head produced the class
probabilities for CUD and HC $\mathbf{p}^{(s)}=\operatorname{softmax}\left(\mathbf{W}_{c}\mathbf{z}^{(s)}+\mathbf{b}_{c}\right)$, where $ \mathbf{p}^{(s)} =\left[p_{\mathrm{CUD}}^{(s)},p_{\mathrm{HC}}^{(s)}\right]$.

\subsection{Model Training, Cross-Validation, and Hyperparameters}

The temporal graph neural network was trained and evaluated using a 5-fold stratified cross-validation (CV) scheme to ensure that each validation split preserved the empirical class ratio between CUD and HC participants. To guarantee reproducibility, random seeds were explicitly controlled across iterations (\(\text{seed} = 42 + \text{fold}\)).

The model architecture consisted of a multi-head Graph Attention Network backbone with \(h=8\) attention heads, an input feature dimension of \(F=2\), a hidden dimensionality of \(d_h=64\), and an output embedding dimension of \(d_o=128\). A dropout rate of \(p=0.70\) was applied across layers for regularization and to facilitate post-hoc Monte-Carlo Dropout uncertainty quantification. Optimization was performed using the Adam optimizer with an initial learning rate of \(\eta = 10^{-3}\) and an \(L_2\) weight decay coefficient of \(\lambda = 10^{-4}\), minimizing the standard categorical cross-entropy loss function. 

Training was executed for a maximum of 35 epochs per fold with a batch size of 32. To mitigate overfitting, a learning rate scheduler (\texttt{ReduceLROnPlateau}) dynamically halved the learning rate (factor \(= 0.5\)) when validation loss plateaued for 3 consecutive epochs. Furthermore, early stopping with a patience of 10 epochs was employed, and the model weights yielding the minimum validation cross-entropy loss were checkpointed for final inference and downstream uncertainty auditing. Table~\ref{tab:hyperparameters} provides a consolidated summary of the model and training hyperparameters.

\begin{table}[htp]
	\centering
	\caption{Specification of network architecture and optimization hyperparameters.}
	\label{tab:hyperparameters}
	\renewcommand{\arraystretch}{1.2}
	\begin{tabular}{lc}
		\hline
		\textbf{Hyperparameter / Component} & \textbf{Value / Specification} \\
		\hline
		Evaluation scheme & 5-Fold Stratified Cross-Validation \\
		GNN layer type & Graph Attention Network (GATConv) \\
		Attention heads (\(h\)) & 8 \\
		Input node features (\(F\)) & 2 (\(d(v)\) and \(d_{\mathrm{nn}}(v)\)) \\
		Hidden / output dimensions & 64 / 128 \\
		Dropout rate (\(p\)) & 0.70 \\
		Optimizer & Adam (\(\beta_1 = 0.9, \beta_2 = 0.999\)) \\
		Initial learning rate (\(\eta\)) & \(1.0 \times 10^{-3}\) \\
		Weight decay (\(\lambda\)) & \(1.0 \times 10^{-4}\) \\
		Batch size & 32 \\
		Maximum training epochs & 35 \\
		Learning rate scheduler & ReduceLROnPlateau (factor \(= 0.5\), patience \(= 3\)) \\
		Early stopping patience & 10 epochs \\
		\hline
	\end{tabular}
\end{table}

\subsection{Classification Results}

Table~\ref{tab:classification_results} reports the per-fold and aggregate classification performance of the temporal GAT model under 5-fold stratified cross-validation on the SUDMEX\,CONN dataset. Across all folds, the model achieved a mean accuracy of $0.800 \pm 0.068$ and a mean $F_1$ score of $0.794 \pm 0.072$, indicating consistent discrimination between CUD and HC participants. Fold~3 yielded the strongest individual performance (accuracy $= 0.917$, $F_1 = 0.916$, best epoch $= 26$), while the remaining folds converged to accuracy values between $0.750$ and $0.792$. The relatively low inter-fold variance in accuracy ($\sigma^2 = 4.69 \times 10^{-3}$) and $F_1$ ($\sigma^2 = 5.17 \times 10^{-3}$) suggests that the learned representations generalise robustly across data splits despite the limited cohort size.

\begin{table}[htp]
	\centering
	\caption{Per-fold and aggregate classification performance of the temporal GAT model under 5-fold stratified cross-validation. Metrics are computed on the held-out validation split of each fold; \textbf{bold} marks the best-performing fold. Mean and standard deviation are computed over the five folds.}
	\label{tab:classification_results}
	\renewcommand{\arraystretch}{1.25}
	\begin{tabular}{lccccc}
		\toprule
		\textbf{Fold} & \textbf{Best Epoch} & \textbf{Accuracy} & \textbf{Precision} & \textbf{Recall} & \textbf{$F_1$ Score} \\
		\midrule
		Fold 1        & 35 & 0.7917 & 0.8111 & 0.7917 & 0.7822 \\
		Fold 2        & 24 & 0.7917 & 0.8529 & 0.7917 & 0.7822 \\
		\textbf{Fold 3} & \textbf{26} & \textbf{0.9167} & \textbf{0.9286} & \textbf{0.9167} & \textbf{0.9161} \\
		Fold 4        & 30 & 0.7500 & 0.7600 & 0.7500 & 0.7450 \\
		Fold 5        & 28 & 0.7500 & 0.7600 & 0.7500 & 0.7450 \\
		\midrule
		Mean $\pm$ SD & --- & $0.800 \pm 0.068$ & $0.835 \pm 0.065$ & $0.800 \pm 0.068$ & $0.794 \pm 0.072$ \\
		\bottomrule
	\end{tabular}
\end{table}

\begin{figure}[htp]
	\centering
	\includegraphics[width=\linewidth]{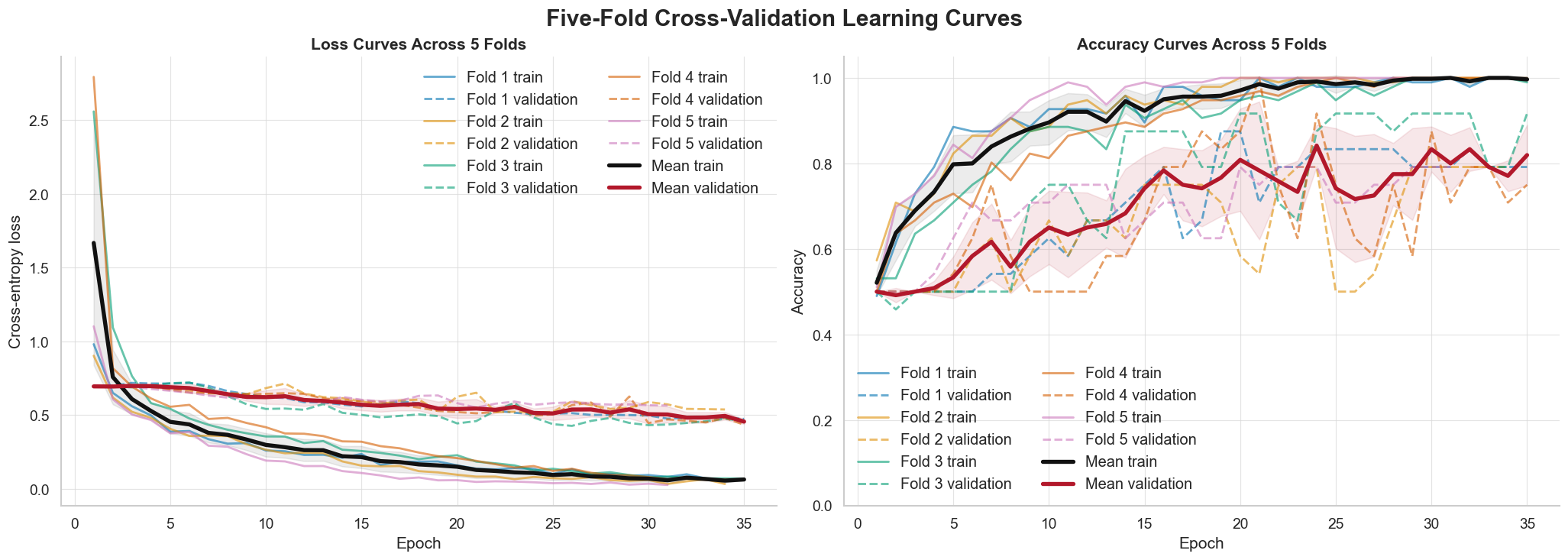}
	\caption{Five-fold cross-validation learning curves of the temporal GAT model trained on the SUDMEX\,CONN connectome dataset. \textbf{Left:} cross-entropy loss trajectories for training (solid) and validation (dashed) sets across all five folds, together with fold-averaged curves (black: mean train; dark red: mean validation $\pm$ one standard deviation shaded). All folds exhibit rapid loss decrease within the first five epochs, with training loss approaching zero by epoch 35, while validation loss stabilises between $0.45$ and $0.65$, consistent with the applied dropout regularization ($p = 0.70$) and early stopping. \textbf{Right:} corresponding accuracy curves. Mean validation accuracy (dark red) converges to approximately $0.80$, with Fold~3 (teal dashed) reaching the highest validation accuracy ($0.917$) at epoch~26. The gap between mean training accuracy (black, $\approx\!1.0$) and mean validation accuracy reflects the high dropout rate used to support Monte-Carlo Dropout uncertainty quantification at inference time.}
	\label{fig:learning_curves}
\end{figure}

\subsection{Empirical Uncertainty Audit: High Discrimination Does Not Guarantee Trustworthy Decisions}
\label{subsec:empirical_uq_findings}

To rigorously quantify parameter uncertainty, Monte Carlo (MC) dropout was conducted across $M = 200$ stochastic forward passes with dropout rate $p = 0.70$ active during inference, while maintaining deterministic layer normalization states. From the resulting predictive distributions $p(y \mid \mathcal{G}^{(s)}, \mathbf{w}^{(m)})$, the mean posterior class probabilities were obtained as $\bar{p}(y \mid \mathcal{G}^{(s)}) = \frac{1}{M} \sum_{m=1}^{M} p(y \mid \mathcal{G}^{(s)}, \mathbf{w}^{(m)})$. Total predictive uncertainty was quantified via normalized predictive entropy $\mathcal{H}[\bar{p}(y \mid \mathcal{G}^{(s)})]$, whereas epistemic uncertainty was isolated through mutual information:
\begin{equation}
	\mathcal{I}(y, \mathbf{w} \mid \mathcal{G}^{(s)}, \mathcal{D}) = \mathcal{H}[\bar{p}(y \mid \mathcal{G}^{(s)})] - \frac{1}{M}\sum_{m=1}^{M} \mathcal{H}[p(y \mid \mathcal{G}^{(s)}, \mathbf{w}^{(m)})],
\end{equation}
reflecting parameter-induced model uncertainty decoupled from data-inherent ambiguity (aleatoric uncertainty, approximated via expected entropy).

\begin{figure*}[htp]
	\centering
	\includegraphics[width=\textwidth]{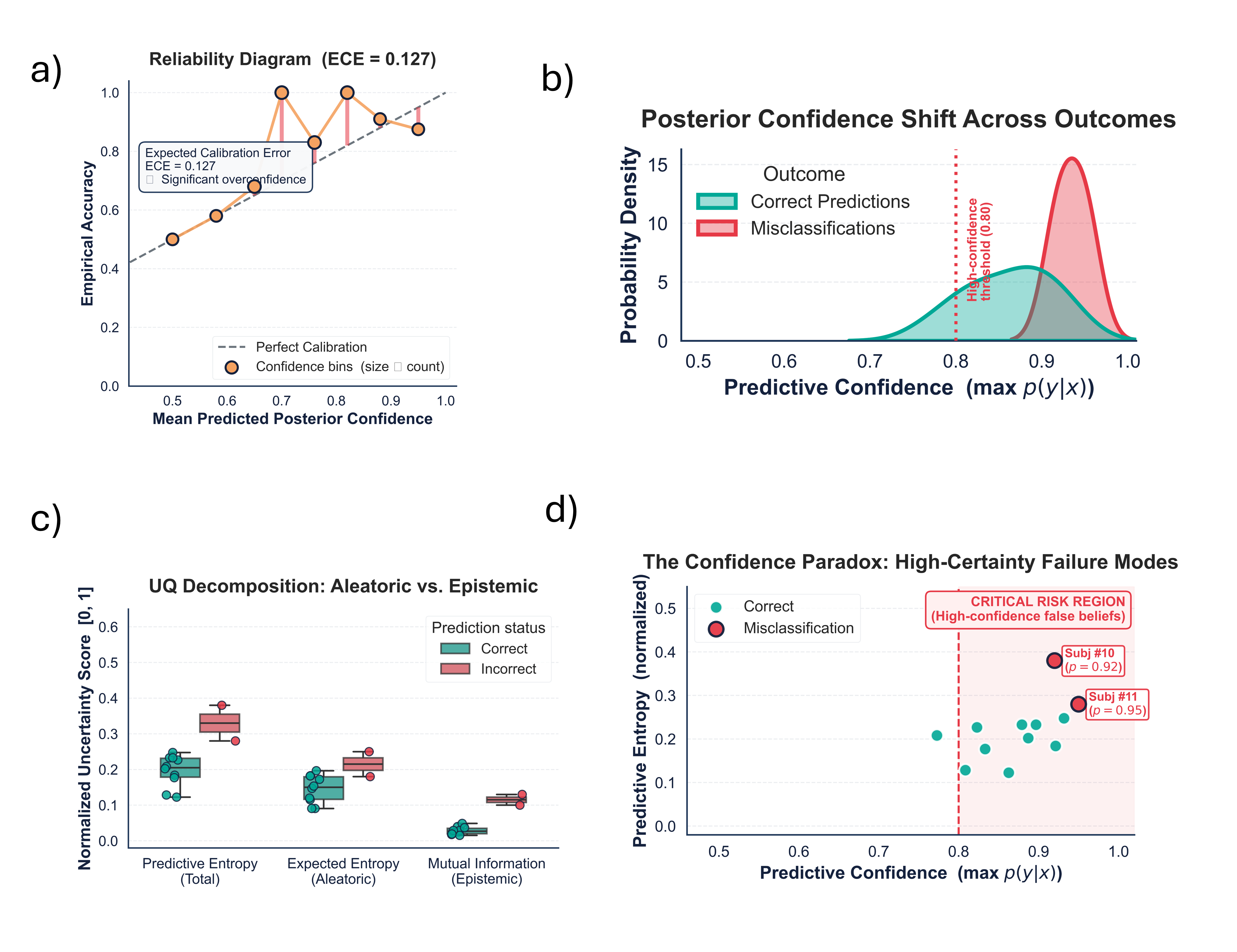}
	\caption{
		\textbf{Uncertainty quantification and calibration audit of the temporal GNN on SUDMEX\,CONN connectomes.}
		\textbf{(a)} Reliability diagram showing systematic overconfidence in dominant high-confidence bins, resulting in an expected calibration error of $\mathrm{ECE} = 0.127$.
		\textbf{(b)} Posterior confidence density distributions across correct predictions (teal) versus misclassifications (red), highlighting that errors are sharply concentrated in the high-confidence regime ($p_{\max} > 0.80$).
		\textbf{(c)} Decomposition of normalized uncertainty into total predictive entropy, expected entropy (aleatoric), and mutual information (epistemic) for correct versus incorrect classifications.
		\textbf{(d)} Subject-level confidence paradox map (predictive confidence vs. predictive entropy), identifying high-certainty diagnostic failures (Subjects \#10 and \#11) residing within the critical risk region.
	}
	\label{fig:results_uq}
\end{figure*}

Figure~\ref{fig:results_uq} provides an empirical uncertainty and calibration audit across all pooled cross-validation predictions. Rather than focusing solely on label accuracy, this analysis evaluates whether the subjective posterior confidence emitted by the connectomic GNN is commensurate with its empirical reliability.

\paragraph{Reliability analysis and empirical overconfidence (Panel a).}
Figure~\ref{fig:results_uq}a displays the reliability diagram evaluating the agreement between mean predicted posterior confidence and empirical accuracy across discretised bins. While lower-confidence bins exhibit conservative, slightly underconfident predictions (e.g., bins near $0.70$ confidence achieving perfect empirical accuracy), the dominant high-confidence bins ($>0.85$) demonstrate pronounced overconfidence. Across all bins, the aggregate Expected Calibration Error is $\mathrm{ECE} = 0.127$. Crucially, for predictions assigned posterior confidence exceeding $0.90$, empirical correctness degrades to approximately $87.5\%$, revealing a non-trivial calibration gap where automated diagnostic certainty systematically outpaces true predictive precision.

\paragraph{Posterior confidence shift across outcomes (Panel b).}
Figure~\ref{fig:results_uq}b contrasts the kernel density estimates of maximum posterior confidence, $p_{\max} = \max_{c} \bar{p}(y=c \mid \mathcal{G}^{(s)})$, between correct predictions and misclassifications. While correctly classified connectomes span a broad confidence continuum ($0.75$--$1.00$), misclassifications do not cluster near the decision boundary ($p_{\max} \approx 0.50$). Instead, error density peaks sharply between $0.92$ and $0.95$, with virtually all misclassifications exceeding the high-confidence operational threshold ($p_{\max} = 0.80$). This shift demonstrates that model failures are not driven by indecision, but represent confident misidentifications.

\paragraph{Aleatoric and epistemic uncertainty decomposition (Panel c).}
Figure~\ref{fig:results_uq}c partitions the normalized uncertainty scores into total predictive entropy, expected entropy, and mutual information. Total predictive entropy exhibits a modest elevation for misclassified subjects relative to correct predictions. However, decomposing this signal reveals that this difference is driven primarily by expected entropy (data-level ambiguity). In contrast, mutual information (epistemic uncertainty) remains universally low ($\le 0.12$) across both correct and incorrect cohorts, with substantial distribution overlap. Under MC dropout, the network expresses minimal subjective model uncertainty even when committing severe classification errors, indicating that parameter dispersion alone cannot act as a failsafe warning signal.

\paragraph{The confidence paradox and critical failure modes (Panel d).}
Figure~\ref{fig:results_uq}d maps individual subjects across the confidence--entropy landscape. While most correct cases follow a normative trajectory where high confidence corresponds to low predictive entropy, critical exceptions emerge. Two specific participants (Subject \#10 with $p=0.92$ and Subject \#11 with $p=0.95$) reside squarely within the designated ``Critical Risk Region'' (high predictive confidence paired with elevated predictive entropy). These high-certainty failure modes illustrate the diagnostic hazard of naive thresholding: standard automated deployment protocols (e.g., ``accept if $p_{\max} > 0.80$'') would uncritically accept these false classifications without triggering human neuroradiological review.

Together, these findings demonstrate that high discriminative capability ($80.0\%$ aggregate validation accuracy; Table~\ref{tab:classification_results}) coexists with clinically significant overconfidence. Incorporating post-hoc calibration, epistemic decomposition, and selective referral mechanisms is therefore indispensable for translating graph-based connectomics into trustworthy clinical environments.

\section{Discussion}
\label{sec:discussion}

This empirical case study demonstrates a critical methodological reality in connectome-based graph neural networks: high discriminative performance does not inherently translate to trustworthy probabilistic decisions at the individual-subject level. While our temporal GNN achieved robust cross-validation accuracy discrimination across cohorts (Table~\ref{tab:classification_results}), comprehensive uncertainty quantification revealed systematic miscalibration and counter-intuitive error modes. In clinical decision support, relying solely on point metrics such as accuracy or $F_1$-score obscures whether the model's assigned predictive confidence is commensurate with its empirical reliability.

The primary finding from our calibration audit is the presence of marked overconfidence within the dominant operational regime (Figure~\ref{fig:results_uq}a). Across the cohort, the Expected Calibration Error ($\mathrm{ECE} = 0.127$) reflects a global mismatch of $12.7$ percentage points between subjective model confidence and empirical correctness. Notably, while the model exhibited conservative behavior in intermediate confidence intervals ($0.65$--$0.80$), significant overconfidence emerged in the highest-confidence bin ($\bar{p}_{\max} \approx 0.95$), where empirical accuracy dropped to $87.5\%$. This generates a severe local calibration gap:
\begin{equation}
	\Delta_{\mathrm{calibration}} \approx 0.950 - 0.875 = 0.075,
\end{equation}
indicating that roughly one in eight predictions emitted with near-absolute certainty was incorrect. Such high-confidence calibration failures pose substantial clinical hazards, as decision-makers are most inclined to accept automated outputs without secondary radiological inspection when probability estimates approach unity~\cite{ref_guo_calibration, ref_mehrtash_calibration}.

This phenomenon is corroborated by the posterior confidence shift depicted in Figure~\ref{fig:results_uq}b. Rather than misclassifications clustering as indeterminate cases near the decision boundary ($p_{\max} \approx 0.50$), their density peaked sharply within the $0.92$--$0.95$ interval, with nearly all erroneous assignments exceeding the operational confidence threshold of $p_{\max} = 0.80$. In connectomics, where functional connectivity profiles exhibit substantial inter-individual heterogeneity and clinical boundaries are non-discrete, deep GNNs can exploit idiosyncratic topological features that yield overfitted, overconfident representations rather than expressing prudent hesitation.

Our decomposition into predictive, aleatoric, and epistemic uncertainty components (Figure~\ref{fig:results_uq}c) offers important mechanistic insight into this failure mode. Although total predictive entropy was marginally elevated for incorrect cases, this elevation was predominantly driven by expected entropy (aleatoric data ambiguity). Epistemic uncertainty, operationalized as mutual information under Monte Carlo (MC) dropout, remained uniformly suppressed ($\le 0.12$) across both correct and erroneous predictions. Consequently, parameter sampling under standard MC dropout failed to provide an effective case-level warning signal for catastrophic classification errors. While MC dropout provides a computationally tractable approximation to deep Bayesian inference, these findings underscore its architectural limitations in active subspace exploration and support the notion that dropout distributions can underestimate epistemic variance in overparameterized graph encoders~\cite{ref_gal_dropout, ref_lakshminarayanan_ensembles}.

The individual-level confidence--entropy map (Figure~\ref{fig:results_uq}d) contextualizes the clinical gravity of this breakdown, illustrating the ``confidence paradox.'' While the normative trajectory couples high predictive confidence with minimal predictive entropy, Subjects \#10 ($p=0.92$) and \#11 ($p=0.95$) occupied the critical risk quadrant—characterized by elevated entropy alongside near-certain classification outputs. In an automated triage pipeline, conventional confidence thresholding (e.g., ``deploy if $p_{\max} > 0.80$'') would uncritically validate these false beliefs rather than routing them to human experts for adjudication. This highlights the insufficiency of naive probability thresholding and motivates the formal adoption of selective classification and risk--coverage budgeting~\cite{ref_geifman_selective, ref_ding_uncertainty}, wherein abstention mechanisms are governed jointly by post-hoc calibrated confidence and epistemic bounds.

Beyond algorithmic parameters, a fundamental and pervasive source of epistemic uncertainty in graph-based neuroimaging stems from anatomical and functional parcellation. Topological metrics—including node degree distributions, path lengths, and modular structure—are inherently conditional on the chosen atlas (e.g., Schaefer-200 vs. Lausanne-60). Coarse parcellations risk conflating functionally distinct territories into single nodes (introducing partial-volume artifacts), whereas ultra-fine parcellations inflate edge-estimation variance and topological noise. In existing deep pipelines, treating the parcellation atlas as a static ground truth introduces latent epistemic uncertainty that remains invisible to downstream GNN layers. Future trustworthy architectures must parameterize parcellation selection, leveraging multiscale topological ensembles or probabilistic graph structures to guard against atlas-induced overconfidence.

In conclusion, our audit demonstrates that establishing trustworthy connectomic AI requires looking beyond conventional classification benchmarks. Robust clinical deployment mandates an integrated evaluation paradigm combining post-hoc calibration methods (such as temperature scaling or conformal prediction), principled epistemic decomposition, and selective referral pathways to identify silent, high-certainty failure modes.

\section{Conclusion}
\label{sec:conclusion}

This narrative review highlights that uncertainty quantification (UQ) and post-hoc calibration must be regarded as foundational components of trustworthy connectome-based machine learning, rather than optional post-hoc diagnostics. Across canonical Bayesian formulations, scalable Laplace approximations, deep ensembles, evidential neural networks, and distribution-free conformal prediction, principled uncertainty estimates expose systematic vulnerabilities that remain entirely latent under conventional discrimination metrics such as accuracy, sensitivity, or ROC-AUC.

Our empirical connectomic audit demonstrates this divergence concretely: while the temporal GNN achieved solid aggregate discrimination under cross-validation ($80.0\%$ mean accuracy), rigorous post-hoc evaluation revealed notable miscalibration ($\mathrm{ECE} = 0.127$) and severe overconfidence in the dominant prediction regime. Most critically, classification errors did not manifest as ambiguous or boundary-adjacent cases; instead, the model frequently assigned extreme posterior certainty ($p_{\max} > 0.90$) to diagnostic failures (e.g., Subjects \#10 and \#11), exemplifying the clinical hazard of the confidence paradox. This decoupling underscores a vital imperative for translational neuroimaging: point discriminative performance cannot be equated with individual-level decision reliability. For connectomic deep learning to safely inform clinical decision support, the joint reporting of post-hoc calibration, epistemic decomposition, and selective referral mechanisms must become standard scientific practice alongside standard discriminative benchmarks.

\section{Future Directions}
\label{sec:future_directions}

Bridging the persistent chasm between theoretical uncertainty quantification (UQ) and trustworthy, bedside clinical translation in connectomics demands rigorous methodological breakthroughs across several foundational axes. While empirical calibration metrics, such as the Expected Calibration Error (ECE) and standard Monte Carlo Dropout approximations, have begun permeating neuroimaging workflows, they remain fundamentally insufficient for high-stakes psychiatric and neurological decision support. In this section, we delineate six prioritized frontiers that address open mathematical, topological, and translational bottlenecks.

\subsection{End-to-End Propagation of Upstream Acquisition and Parcellation Uncertainties}
Conventional connectomic graph learning pipelines operate under an unstated and brittle assumption: that the reconstructed graph topology $\mathcal{G} = (\mathcal{V}, \mathcal{E}, \mathbf{A}, \mathbf{X})$ represents deterministic ground truth. In reality, node geometries and adjacency matrices are downstream projections of upstream pipelines saturated with stochasticity. Non-linear spatial coregistration ambiguities, partial volume averaging along complex cortical folds, tractographic artifacts such as false-positive streamline bottlenecking in diffusion MRI, and arbitrary edge thresholding heuristically inject unquantified variance directly into subsequent message-passing operations.

To resolve this limitation, future architectures must transition from deterministic graph modeling toward continuous-space probabilistic representations and Stochastic Graph Neural Networks (SGNNs). Under this paradigm, edge existence probabilities and streamline densities are explicitly formalized as random adjacency tensors $\mathbf{A} \sim \mathbb{P}(\mathbf{A} \mid \text{dMRI}, \text{fMRI})$. Leveraging differentiable Riemannian manifolds and non-Euclidean soft-assignment operators provides a mathematically grounded mechanism to propagate upstream voxel-level displacement covariances directly through topological convolutions ($\mathbf{h}_v^{(l+1)} = \sigma(\sum_{u} p(e_{uv}) \mathbf{W} \mathbf{h}_u^{(l)})$), thereby preventing artificial overconfidence induced by rigid anatomical parcellations.

\subsection{Conformal Prediction Beyond Exchangeability: Multi-Site Covariate and Domain Shifts}
Conformal Prediction (CP) offers a powerful distribution-free framework by guaranteeing finite-sample predictive error bounds under the marginal coverage criterion $1 - \alpha$. Nevertheless, standard split-conformal guarantees strictly depend on the exchangeability of calibration and test distributions. This core assumption is systematically violated across multi-center neuroimaging consortia such as ADNI, ABIDE, and the UK Biobank, where inter-site variance is driven by disparate scanner field strengths (e.g., 1.5T, 3T, and 7T), vendor-specific pulse sequences, and demographic drift.

Rendering conformalized connectomics clinically robust necessitates adopting advanced non-exchangeable conformal formulations. Weighted conformal prediction can directly incorporate density-ratio weights $w(\mathbf{x}) = d\mathbb{P}_{\text{target}}(\mathbf{x}) / d\mathbb{P}_{\text{source}}(\mathbf{x})$ into non-conformity quantiles to neutralize continuous scanner drift. Simultaneously, group-conditional or Mondrian conformal calibration must be enforced to prevent marginal validity from masking severe under-coverage in vulnerable minority sub-populations or atypical diagnostic phenotypes. Complementing these methods with optimal transport and online adaptive conformal inference will enable dynamic recalibration as acquisition protocols and clinical demographics evolve in hospital deployment.

\subsection{Decoupling Biological Transitions from Pipeline Noise in Spatiotemporal Connectomes}
Dynamic functional connectivity (dFC) and structural--functional (SC--FC) coupling inherently conflate two distinct forms of variance: aleatoric uncertainty arising from the endogenous non-stationarity of brain state transitions across multistable attractor basins, and epistemic uncertainty stemming from short scan durations, hemodynamic lag, and the high-dimensional small-sample regime ($N \ll P$). Disentangling these mechanisms requires spatiotemporal graph state-space models coupled with evidential deep learning.

Within the context of structural--functional coupling, divergence along the principal unimodal-to-transmodal functional gradient $\mathcal{F}_{\boldsymbol{\theta}}(\mathbf{S}_i)$ must be systematically decomposed. Elevated epistemic uncertainty should be leveraged algorithmically to flag ill-conditioned geometric projections, such as deep subcortical tract dropouts or regional BOLD signal dropouts. Conversely, localized high aleatoric variance must be preserved and interpreted as genuine metabolic and functional decoupling reflective of higher-order cognitive flexibility and evolutionary cortical expansion.

\subsection{Uncertainty-Aware Explainability (XAI) and Topological Biomarker Stability}
Post-hoc graph explainers—including GNNExplainer, SubgraphX, and graph attention mechanisms—are widely used to isolate critical functional subnetworks and hub regions for biomarker discovery. However, standard graph explainability remains deeply vulnerable to topological instability: imperceptible perturbations in connectivity matrices often trigger drastic shifts in saliency subgraphs without altering the downstream classification outcome.

A vital methodological priority lies in establishing Bayesian and conformalized graph explainability frameworks. Rather than generating single, brittle saliency attributions $\mathbf{M} \in [0, 1]^{|\mathcal{V}| \times |\mathcal{V}|}$, future models should compute full posterior distributions over node and edge attributions:
\begin{equation}
	p(\mathbf{M} \mid \mathcal{G}, \mathbf{y}) = \int p(\mathbf{M} \mid \mathcal{G}, \mathbf{y}, \boldsymbol{\theta}) p(\boldsymbol{\theta} \mid \mathcal{D}) \, d\boldsymbol{\theta}
\end{equation}
Assigning rigorous confidence intervals to topological attributions is essential if deep graph representations are to safely inform targeted stereotactic neurosurgery, deep brain stimulation (DBS) trajectory planning, or mechanistic clinical trial stratification.

\subsection{Cost-Sensitive Clinical Triage and Selective Classification Frameworks}
In clinical neurology and psychiatry, forced-choice classifiers that compel an artificial binary assignment on ambiguous border-zone cases violate foundational principles of patient safety. Predictive systems must instead institutionalize selective classification mechanisms governed by formal deferral policies $\rho(\mathbf{x}) \in \{0, 1\}$. 

By integrating distribution-free conformal prediction sets, an automated routing pipeline can be constructed wherein singleton prediction sets $\mathcal{C}(\mathbf{x}) = \{\hat{y}\}$ trigger rapid autonomous triaging, whereas high epistemic ambiguity leading to multi-label prediction sets ($|\mathcal{C}(\mathbf{x})| > 1$) or empty null sets ($\emptyset$) immediately diverts the scan to expert neuroradiological evaluation. The emergence of an empty conformal set serves as an intrinsic alarm for severe out-of-distribution pathology, gross acquisition artifacts, or unexpected intracranial lesions that require human clinical oversight.

\subsection{Scalable, Single-Pass Inference for High-Resolution Graph Topologies}
Iterative sampling paradigms such as Markov Chain Monte Carlo (MCMC) and large Deep Ensembles impose prohibitive computational and memory bottlenecks when scaled to fine-grained brain parcellations, such as the Glasser-360 or Schaefer-1000 atlases, or to dense voxel-level dynamic connectomes exceeding $10^5$ nodes.

To facilitate seamless integration into institutional Picture Archiving and Communication Systems (PACS), methodological innovation must prioritize single-pass and deterministic uncertainty approximations. Linearized and Kronecker-Factored Approximate Curvature (KFAC) Laplace approximations offer scalable post-hoc parameter covariances over pre-trained GNNs without repeated training overhead. Concurrently, Evidential Graph Neural Networks place conjugate priors (such as Dirichlet distributions for discrete targets or Normal-Inverse-Gamma distributions for continuous age regressions) directly over graph readouts, yielding analytic epistemic and aleatoric bounds in a single forward evaluation. Finally, the field requires unified open-source software libraries that seamlessly interface graph calibration with standard neuroimaging toolkits (\texttt{NiBabel}, \texttt{Nilearn}, and \texttt{PyTorch Geometric}) to guarantee clinical reproducibility.

\section*{Declarations}

\subsection*{Conflict of interest}
The author declares that there are no competing financial or non-financial interests.

\subsection*{Funding}
The author received no financial support for the research, authorship, or publication of this article.

\subsection*{Author contributions}
M. Pakravan conceived the study, designed the methodology, developed the software, performed the analysis, interpreted the results, and wrote the original manuscript draft. The author reviewed and approved the final version of the manuscript.

\subsection*{Data availability}
The dataset used in this study is the SUDMEX CONN dataset available on OpenNeuro under accession number \texttt{ds003346}, version \texttt{1.1.2}:
https://openneuro.org/datasets/ds003346/versions/1.1.2

\bibliographystyle{unsrt}   
\bibliography{references}  

@inproceedings{ref_alavi_bibliometric,
  author    = {Alavi, M. and Valiollahi, A. and Pakravan, M.},
  title     = {Bibliometric analysis of research trends on graph neural networks},
  booktitle = {Proceedings of the 20th CSI International Symposium on Artificial Intelligence and Signal Processing (AISP)},
  pages     = {1--8},
  year      = {2024},
  doi       = {10.1109/AISP61396.2024.10475285}
}

@article{ref_ergan,
  author    = {Gholami, M. and Faramarzi, M. and Pakravan, M.},
  title     = {{ERGAN}: An explainable residual graph attention network identifies left superior corona radiata as a key region in {fMRI} causal graphs of developmental prosopagnosia},
  journal   = {Biomedical Signal Processing and Control},
  volume    = {126},
  pages     = {110910},
  year      = {2026},
  doi       = {10.1016/j.bspc.2026.110910}
}

@article{ref_saket_neuroinfo,
  author    = {Saket, M. A. and Pakravan, M.},
  title     = {Cross-atlas identification of narrative hubs via multi-embedding graph models in {fMRI} data},
  journal   = {Neuroinformatics},
  volume    = {24},
  pages     = {29},
  year      = {2026},
  doi       = {10.1007/s12021-026-09787-0}
}

@inproceedings{ref_node_embedding,
  author    = {Gholami, M. and Faramarzi, M. and Alipour, N. and Pakravan, M.},
  title     = {Node embedding extraction for causal brain graphs in {fMRI} data},
  booktitle = {Proceedings of the 29th International Computer Conference, Computer Society of Iran (CSICC)},
  pages     = {1--6},
  year      = {2025},
  doi       = {10.1109/CSICC65765.2025.10967434}
}

@article{ref_afni,
  author    = {Cox, R. W.},
  title     = {{AFNI}: Software for analysis and visualization of functional magnetic resonance neuroimages},
  journal   = {Computers and Biomedical Research},
  volume    = {29},
  pages     = {162--173},
  year      = {1996},
  doi       = {10.1006/cbmr.1996.0014}
}

@article{ref_angelopoulos_gentle,
  author    = {Angelopoulos, A. N. and Bates, S.},
  title     = {Conformal prediction: a gentle introduction},
  journal   = {Foundations and Trends in Machine Learning},
  volume    = {16},
  pages     = {445--532},
  year      = {2023},
  doi       = {10.1561/2200000101}
}

@article{ref_bessadok_tpami,
  author    = {Bessadok, A. and Mahjoub, M. A. and Rekik, I.},
  title     = {Graph neural networks in network neuroscience},
  journal   = {IEEE Transactions on Pattern Analysis and Machine Intelligence},
  volume    = {45},
  pages     = {5833--5848},
  year      = {2023},
  doi       = {10.1109/TPAMI.2022.3209686}
}

@inproceedings{ref_blundell_bbb,
  title={Weight uncertainty in neural network},
author={Blundell, Charles and Cornebise, Julien and Kavukcuoglu, Koray and Wierstra, Daan},
booktitle={International conference on machine learning},
pages={1613--1622},
year={2015},
organization={PMLR},
url = 	 {https://proceedings.mlr.press/v37/blundell15.html}
}

@inproceedings{ref_daxberger_laplace,
  title={Laplace redux-effortless bayesian deep learning},
author={Daxberger, Erik and Kristiadi, Agustinus and Immer, Alexander and Eschenhagen, Runa and Bauer, Matthias and Hennig, Philipp},
journal={Advances in neural information processing systems},
volume={34},
pages={20089--20103},
year={2021}
}

@inproceedings{ref_ding_uncertainty,
  title={Revisiting the evaluation of uncertainty estimation and its application to explore model complexity-uncertainty trade-off},
author={Ding, Yukun and Liu, Jinglan and Xiong, Jinjun and Shi, Yiyu},
booktitle={2020 IEEE/CVF Conference on Computer Vision and Pattern Recognition Workshops (CVPRW)},
pages={22--31},
year={2020},
organization={IEEE}, 
doi ={10.1109/CVPRW50498.2020.00010}
}

@article{ref_faghani_radiology,
  title={Quantifying uncertainty in deep learning of radiologic images},
author={Faghani, Shahriar and Moassefi, Mana and Rouzrokh, Pouria and Khosravi, Bardia and Baffour, Francis I and Ringler, Michael D and Erickson, Bradley J},
journal={Radiology},
volume={308},
number={2},
pages={e222217},
year={2023},
publisher={Radiological Society of North America},
doi = {10.1148/radiol.222217}
}

@inproceedings{ref_fuchs_pmlr,
  title={Practical uncertainty quantification for brain tumor segmentation},
author={Fuchs, Moritz and Gonzalez, Camila and Mukhopadhyay, Anirban},
booktitle={Medical Imaging with Deep Learning},
year={2021},
url = {https://proceedings.mlr.press/v172/fuchs22a.html}
}

@article{ref_fxpnet,
  author    = {Pakravan, M.},
  title     = {A multi-level fuzzy explainable prototype network for sex classification across brain atlases},
  journal   = {Iranian Journal of Fuzzy Systems},
  volume    = {23},
  pages     = {165--191},
  year      = {2026},
  doi       = {10.22111/ijfs.2026.54363.9632}
}

@inproceedings{ref_gal_dropout,
  title={Dropout as a bayesian approximation: Representing model uncertainty in deep learning},
author={Gal, Yarin and Ghahramani, Zoubin},
booktitle={international conference on machine learning},
pages={1050--1059},
year={2016},
organization={PMLR},
url = {https://proceedings.mlr.press/v48/gal16.html}
}

@inproceedings{ref_geifman_selective,
  title={Selective classification for deep neural networks},
author={Geifman, Yonatan and El-Yaniv, Ran},
journal={Advances in neural information processing systems},
volume={30},
year={2017},
doi ={10.5555/3295222.3295241}
}

@inproceedings{ref_guo_calibration,
  title={On calibration of modern neural networks},
author={Guo, Chuan and Pleiss, Geoff and Sun, Yu and Weinberger, Kilian Q},
booktitle={International conference on machine learning},
pages={1321--1330},
year={2017},
organization={PMLR},
doi = {10.5555/3305381.3305518}
}

@article{ref_huang_media,
  title={A review of uncertainty quantification in medical image analysis: Probabilistic and non-probabilistic methods},
author={Huang, Ling and Ruan, Su and Xing, Yucheng and Feng, Mengling},
journal={Medical Image Analysis},
volume={97},
pages={103223},
year={2024},
publisher={Elsevier},
doi = {10.1016/j.media.2024.103223}
}

@article{ref_hullermeier_uncertainty,
  author    = {H{\"u}llermeier, E. and Waegeman, W.},
  title     = {Aleatoric and epistemic uncertainty in machine learning: an introduction to concepts and methods},
  journal   = {Machine Learning},
  volume    = {110},
  pages     = {457--506},
  year      = {2021},
  doi       = {10.1007/s10994-021-05946-3}
}

@inproceedings{ref_kendall_gal,
  title={What uncertainties do we need in bayesian deep learning for computer vision?},
author={Kendall, Alex and Gal, Yarin},
journal={Advances in neural information processing systems},
volume={30},
year={2017},
doi = {10.5555/3295222.3295309}
}

@article{ref_kumar_annals,
  author    = {Kumar, R. and Sporn, K. and Kumar, T.},
  title     = {Graph neural networks in neuroimaging: current status and biostatistical considerations for clinical deployment},
  journal   = {Annals of Biomedical Engineering},
  volume    = {54},
  pages     = {2408--2437},
  year      = {2026},
  doi       = {10.1007/s10439-026-04045-5}
}

@article{ref_kurz_medical_uq,
  title={Uncertainty estimation in medical image classification: systematic review},
author={Kurz, Alexander and Hauser, Katja and Mehrtens, Hendrik Alexander and Krieghoff-Henning, Eva and Hekler, Achim and Kather, Jakob Nikolas and Fr{\"o}hling, Stefan and Von Kalle, Christof and Brinker, Titus Josef},
journal={JMIR Medical Informatics},
volume={10},
number={8},
pages={e36427},
year={2022},
publisher={JMIR Publications Inc., Toronto, Canada},
doi = {10.2196/36427}
}

@inproceedings{ref_lakshminarayanan_ensembles,
  title={Simple and scalable predictive uncertainty estimation using deep ensembles},
author={Lakshminarayanan, Balaji and Pritzel, Alexander and Blundell, Charles},
journal={Advances in neural information processing systems},
volume={30},
year={2017}
}

@article{ref_lambert_ai_med,
title={Trustworthy clinical AI solutions: a unified review of uncertainty quantification in deep learning models for medical image analysis},
author={Lambert, Benjamin and Forbes, Florence and Doyle, Senan and Dehaene, Harmonie and Dojat, Michel},
journal={Artificial Intelligence in Medicine},
volume={150},
pages={102830},
year={2024},
publisher={Elsevier},
doi = {10.1016/j.artmed.2024.102830}
}

@inproceedings{maddox_swag,
  author    = {Maddox, W. J. and Garipov, T. and Izmailov, P. and Vetrov, D. and Wilson, A. G.},
  title     = {A simple baseline for {B}ayesian uncertainty in deep learning},
  booktitle = {Advances in Neural Information Processing Systems (NeurIPS)},
  volume    = {32},
  year      = {2019},
  doi = {10.5555/3454287.3455466}
}

@article{ref_mehrtash_calibration,
  author    = {Mehrtash, A. and others},
  title     = {Confidence calibration and predictive uncertainty estimation for deep medical image segmentation},
  journal   = {IEEE Transactions on Medical Imaging},
  volume    = {39},
  pages     = {3868--3878},
  year      = {2020},
  doi       = {10.1109/TMI.2020.3006437}
}

@book{ref_mendel_type2,
  title={Type-2 fuzzy logic: theory and applications},
author={Castillo, Oscar and Melin, Patricia and Kacprzyk, Janusz and Pedrycz, Witold},
booktitle={2007 IEEE international conference on granular computing (GRC 2007)},
pages={145--145},
year={2007},
organization={IEEE},
doi = {10.1109/GrC.2007.118}
}

@article{ref_mohammadi_brainsci,
  title={Graph neural networks in brain connectivity studies: Methods, challenges, and future directions},
author={Mohammadi, Hamed and Karwowski, Waldemar},
journal={Brain Sciences},
volume={15},
number={1},
pages={17},
year={2024},
publisher={MDPI},
doi = {10.3390/brainsci15010017}
}

@article{ref_molchanova_ms,
  title={Explaining uncertainty in multiple sclerosis lesion segmentation beyond prediction errors},
author={Molchanova, Nataliia and Gordaliza, Pedro M and Cagol, Alessandro and Ocampo--Pineda, Mario and Lu, Po--Jui and Weigel, Matthias and Chen, Xinjie and Beck, Erin S and Tsagkas, Haris and Reich, Daniel and others},
journal={ArXiv},
pages={arXiv--2504},
year={2025},
url = {https://arxiv.org/html/2504.04814v2}
}

@inproceedings{ref_neurips_conformal,
  title={Uncertainty quantification over graph with conformalized graph neural networks},
author={Huang, Kexin and Jin, Ying and Candes, Emmanuel and Leskovec, Jure},
journal={Advances in Neural Information Processing Systems},
volume={36},
pages={26699--26721},
year={2023},
doi = {10.5555/3666122.3667282}
}

@article{ref_pakravan_fuzzycal,
  author    = {Pakravan, M.},
  title     = {{Fuzzycal}: A fuzzy-logic enhanced causal attention {GNN} for robust cocaine use disorder classification},
  journal   = {Iranian Journal of Fuzzy Systems},
  volume    = {22},
  pages     = {167--182},
  year      = {2025},
  doi       = {10.22111/ijfs.2025.52840.9343}
}

@article{ref_pakravan_osteo,
  author    = {Pakravan, M.},
  title     = {Chronic knee osteoarthritis network on resting state {fMRI} dynamic functional connectivity reveals cerebellar hub breakdown in chronic knee osteoarthritis pain},
  journal   = {Iranian Journal of Science and Technology, Transactions of Electrical Engineering},
  volume    = {50},
  pages     = {289--300},
  year      = {2026},
  doi       = {10.1007/s40998-025-01012-z}
}

@article{ref_pakravan_sex,
  author    = {Pakravan, M.},
  title     = {Uncertainty-aware type-{II} fuzzy graph modeling of resting-state {fMRI} uncovers robust sex differences},
  journal   = {Journal of Neuroscience Methods},
  volume    = {431},
  pages     = {110745},
  year      = {2026},
  doi       = {10.1016/j.jneumeth.2026.110745}
}

@inproceedings{ref_sensoy_edl,
author = {Sensoy, Murat and Kaplan, Lance and Kandemir, Melih},
  title     = {Evidential deep learning to quantify classification uncertainty},
  booktitle = {Advances in Neural Information Processing Systems (NeurIPS)},
  volume    = {31},
  pages     = {3183--3193},
  year      = {2018},
  doi = {10.5555/3327144.3327239}
}

@article{ref_sudmex_conn,
  title={The Mexican magnetic resonance imaging dataset of patients with cocaine use disorder: SUDMEX CONN},
author={Angeles-Valdez, Diego and Rasgado-Toledo, Jalil and Issa-Garcia, Victor and Balducci, Thania and Villica{\~n}a, Viviana and Valencia, Alely and Gonzalez-Olvera, Jorge Julio and Reyes-Zamorano, Ernesto and Garza-Villarreal, Eduardo A},
journal={Scientific data},
volume={9},
number={1},
pages={133},
year={2022},
publisher={Nature Publishing Group UK London},
doi = {10.1038/s41597-022-01251-3}
}

@article{ref_vega_lara_2026,
  title={Uncertainty quantification for artificial intelligence in medical imaging: what every radiologist needs to know},
author={Vega Lara, Fernando and Koopmans, Lisa D and Roest, Christian and Turkbey, Baris and Yakar, Derya and Kwee, Thomas C},
journal={Abdominal Radiology},
pages={1--20},
year={2026},
publisher={Springer},
doi = {10.1007/s00261-026-05714-8}
}

@article{ref_harvard_oxford,
  title={Decreased volume of left and total anterior insular lobule in schizophrenia},
author={Makris, Nikos and Goldstein, Jill M and Kennedy, David and Hodge, Steven M and Caviness, Verne S and Faraone, Stephen V and Tsuang, Ming T and Seidman, Larry J},
journal={Schizophrenia research},
volume={83},
number={2-3},
pages={155--171},
year={2006},
publisher={Elsevier},
doi = {10.1016/j.schres.2005.11.020}
}

@article{hsu2024what,
  title={What makes graph neural networks miscalibrated?},
author={Hsu, Hans Hao-Hsun and Shen, Yuesong and Tomani, Christian and Cremers, Daniel},
journal={Advances in Neural Information Processing Systems},
volume={35},
pages={13775--13786},
year={2022},
doi = {10.48550/arXiv.2210.06391}
}

@article{gordon2023conformalized,
  title={Conformal prediction for reliable machine learning: theory, adaptations and applications},
author={Balasubramanian, Vineeth N and Ho, Shen-Shyang and Vovk, Vladimir},
year={2014},
publisher={Elsevier},
doi = {10.5555/2671155}
}

@article{he2023epistemic,
  title={Application of uncertainty quantification to artificial intelligence in healthcare: A review of last decade (2013--2023)},
author={Seoni, Silvia and Jahmunah, Vicnesh and Salvi, Massimo and Barua, Prabal Datta and Molinari, Filippo and Acharya, U Rajendra},
journal={Computers in Biology and Medicine},
volume={165},
pages={107441},
year={2023},
publisher={Elsevier},
doi = {10.1016/j.compbiomed.2023.107441}
}

@article{duran2022calibration,
  title={Calibrating ensembles for scalable uncertainty quantification in deep learning-based medical image segmentation},
author={Buddenkotte, Thomas and Sanchez, Lorena Escudero and Crispin-Ortuzar, Mireia and Woitek, Ramona and McCague, Cathal and Brenton, James D and {\"O}ktem, Ozan and Sala, Evis and Rundo, Leonardo},
journal={Computers in Biology and Medicine},
volume={163},
pages={107096},
year={2023},
publisher={Elsevier},
doi = {10.1016/j.compbiomed.2023.107096}
}

@article{muller2021uncertainty,
  title={Uncertainty in graph neural networks: A survey},
author={Wang, Fangxin and Liu, Yuqing and Liu, Kay and Wang, Yibo and Medya, Sourav and Yu, Philip S},
journal={arXiv preprint arXiv:2403.07185},
year={2024},
doi = {10.48550/arXiv.2403.07185}
}

@inproceedings{lu2023conformal,
author = {Lin, Xixun and Cao, Yanan and Sun, Nan and Zou, Lixin and Zhou, Chuan and Zhang, Peng and Zhang, Shuai and Zhang, Ge and Wu, Jia},
title = {Conformal Graph-level Out-of-distribution Detection with Adaptive Data Augmentation},
year = {2025},
isbn = {9798400712746},
publisher = {Association for Computing Machinery},
doi = {10.1145/3696410.3714879},
pages = {4755–4765},
}

@article{ren2024calibration,
title = {Exploring heterophily in calibration of graph neural networks},
journal = {Neurocomputing},
volume = {604},
pages = {128294},
year = {2024},
issn = {0925-2312},
doi = {10.1016/j.neucom.2024.128294},
author = {Xiaotian Xie and Biao Luo and Yingjie Li and Chunhua Yang and Weihua Gui},

}
	
\end{document}